\documentclass{article}

\usepackage{arxiv}
\usepackage[utf8]{inputenc} 
\usepackage[T1]{fontenc}    
\usepackage{url}            
\usepackage{booktabs}       
\usepackage{amsmath}
\usepackage{amsfonts}       
\usepackage{nicefrac}       
\usepackage{microtype}      
\usepackage{lipsum}
\usepackage{graphicx}
\usepackage{comment}
\usepackage{xcolor}
\usepackage{longtable}
\usepackage{float}
\usepackage{amsmath}
\usepackage{graphicx}

\newcommand{\mb}{\mathbf}

\graphicspath{{figs/}}
\title{Edge-similarity-aware Graph Neural Networks }

\author{
  Vincent Mallet \\
  Pasteur Institute \\
  Les Mines-Paristech \\
  \texttt{vincent.mallet96@gmail.com}
  \And
    Carlos Oliver  \\
  Department of Computer Science, McGill University \\ 
  Montreal Institute for Learning Algorithms\\
  \texttt{carlos.gonzalezoliver@mail.mcgill.ca}
  \And 
  William L. Hamilton \\
  Department of Computer Science, McGill University \\ 
  Montreal Institute for Learning Algorithms\\
  \texttt{wlh@cs.mcgill.ca}
}

\begin{document}

\maketitle

\noindent\hspace{0.15\linewidth}\begin{minipage}{0.7\textwidth}

\section*{Abstract}
Graph are a ubiquitous data representation, as they represent a flexible and compact representation.
For instance, the 3D structure of RNA can be efficiently represented as \textit{2.5D graphs}, graphs whose nodes are nucleotides and edges represent chemical interactions.
In this setting, we have biological evidence of the similarity between the edge types, as some chemical interactions are more similar than others.

Machine learning on graphs have recently experienced a breakthrough with the introduction of Graph Neural Networks.
This algorithm can be framed as a message passing algorithm between graph nodes over graph edges.
These messages can depend on the edge type they are transmitted through, but no method currently constrains how a message is altered when the edge type changes.

Motivated by the RNA use case, in this project we introduce a graph neural network layer which can leverage prior information about similarities between edges.
We show that despite the theoretical appeal of including this similarity prior, the empirical performance is not enhanced on the tasks and datasets we include here. \\
\end{minipage}

\section{Introduction}

\subsection{Graph neural networks}
Machine learning on graphs has experienced a breakthrough with the introduction of graph convolutional networks \cite{kipf2016semi}.
A line of papers has made the formulation evolve towards a message passing algorithms. 
All nodes start with an initial feature vector and at each layer, they convolve their feature vector with a parameter tensor and send it to their neighbors.
Then these messages get aggregated, possibly with an attention mechanism \cite{gat}.

\subsection{Relation types}
This approach was very successful to learn on graphs for several domains such as biology or social science.
Open Graph Benchmark \cite{hu2020open} is an initiative to goup several graph data sets into a unified benchmark.
Different relation types exist in several data sets such as knowledge graphs, protein networks, molecules or drug interaction networks.
The possibility to use different convolution kernels depending on the edge type that link two nodes was introduced by a model know as the relational graph convolutional network \cite{rgcn}.
Later, some papers added the possibility to also add a relation-dependent attention mechanism \cite{busbridge2019relational, wang2020relational} while others tried replacing the relation-dependent convolutional kernels with relation-dependent scaling coefficients \cite{sacn}.
Overall, it seems that none of these attention-like mechanisms significantly improved the predictive performances.
Finally, other models introduced new message passing framework that
include edge-type embeddings in the message passing procedure, which results in improved performances \cite{vashishth2019composition}.
However, none of these methods include prior information about the relation-types similarities.

\subsection{RNA 2.5D Graphs and Edge-type Prior Information}
To capture the tertiary structure of RNA in a computationally feasible manner, a growing number of algorithms make use of 2.5D graph networks, whose nodes are nucleotides and whose edge types are structural categories of interactions \cite{mallet2021rnaglib}.
This representation carries a strong prior and has shown to be efficient for machine learning.
Moreover, in this RNA setting, we have an additional prior information about the different edge types.
Indeed, the edges represent structural categories of interactions: how the nucleotides spatially interact with each other.
Therefore, some pairs of categories are more similar than others, and these similarities are known and denoted as isostericity \cite{stombaugh2009frequency}.

\subsection{Contribution} 
In this paper, we introduce a new graph neural network layer to leverage the isostericity prior information.
It falls into previous frameworks but to have this prior information constraint, we have to use specific instances of these frameworks and to add a specific loss.
We introduce the mathematical framing that motivates us, implement it and investigate its performance on several RNA tasks.
In this setting, these new layers don't seem to result in increased performances. The code is available at \texttt{https://github.com/Vincentx15/RNAttentional}.

\section{Methods}

\subsection{Initial Setup and Mathematical Framing}

Following \cite{busbridge2019relational, wang2020relational}, we define a general layer update as:

\begin{align*}
    \mb{h}_v^{(t)} = \textrm{ReLU}\left((\mb{W}^{(t)}_{0}\mb{h}_v^{(t-1)} + \sum_{r \in R}\sum_{u \in \mathcal{N}_r(v)}\mb{\alpha_r} \mb{W}_r\mb{h}_u^{(t-1)}\right),
\end{align*}

Where $\mb{W}_r$ is a propagation matrix specific to edge-type and $\mb{\alpha_r}$ are attention weights. They can be either a constant equal to one \cite{kipf2016semi}, a relation-dependent constant \cite{sacn}, usual multiplicative or additive attention weights \cite{busbridge2019relational} or the output of a small MLP \cite{wang2020relational}.

Moreover the basis sharing trick consists in generating the relational propagation matrices $\mb{W}_r$ from a smaller set of basis and coefficients to add regularization by adding a rank constraint :
\begin{align*}
\mb{W}_{r} &= \sum_b a_{r,b}\mb{V}_b\\
\end{align*}

We are now interested in having a network that conveys similar messages through similar edges, as an alternative way to constrain the convolutional kernel.
This would enable baking in the isostericity values into the network.

Let us denote as $\Delta(r_1, r_2)$ the norm of the difference of the messages incurred by swapping an edge type for another in a graph.
We have : 

\begin{align*}
\Delta(r_1, r_2) = \| \mb{\alpha_{r_1}} \mb{W}_{r_1}\mb{h}_u - \mb{\alpha_{r_2}} \mb{W}_{r_2}\mb{h}_u\|^2.
\end{align*}

Let $Iso(r_1, r_2)$ be the isostericity value between two edges. 
The goal is to have :
\begin{align*}
\Delta(r_1, r_2) \sim Iso(r_1, r_2)
\end{align*}

\subsection{Enforcing Edge Similarity}

For arbitrary $\mb{\alpha_{r}}\mb{W}_{r}$, we only have :
\begin{align*}
\Delta(r_1, r_2) \leq \| \mb{\alpha_{r_1}} \mb{W}_{r_1} - \mb{\alpha_{r_2}} \mb{W}_{r_2} \|^2 \|\mb{h}_u\|^2
\end{align*}
We see here that there is no real control over how the message changes for dissimilar relationships, that is the problem with controlling using operator norm.
If we push $\|\mb{W}_{r_1} - \mb{W}_{r_2}\| $ to a value that is not zero, which will be the case for the isostericity, we have no control onto the message alteration.

The only way to get control over this norm is to use a scaling that does not depend on the message dimension, namely attention.
However we cannot have several tensor weights for scaling the input messages.
We then have : 
\begin{align*}
\| \alpha_{r_1}\mb{W} \mb{h}_u - \alpha_{r_2} \mb{W} \mb{h}_u\| 
= \| \alpha_{r_1} - \alpha_{r_2} \|  \| \mb{W} \mb{h}_u \|.
\end{align*}

Let us assume now that we have $
\| \alpha_{r_1} - \alpha_{r_2} \| \sim Iso(r_1, r_2)$, with the attention relationship, we have : 
\begin{align*}
\frac{\Delta(r_1, r_2)}
{\Delta(r_1, r_3)} =
\frac{\| \alpha_{r_1} - \alpha_{r_2} \|}
{\| \alpha_{r_1} - \alpha_{r_3} \|}
\sim 
\frac{Iso(r_1, r_2)}
{Iso(r_1, r_3)},
\end{align*}

which is a control over how messages change when they are convoluted over different edge types.
Therefore, enforcing our prior into the network boils down to enforcing the following relationship : $
\| \alpha_{r_1} - \alpha_{r_2} \| \sim Iso(r_1, r_2)$

To do so, during forward propagation, we randomly sample a subset of edges $\mathcal{S}$ and randomly affect them another edge-type, $\sigma(e)$.
Then we add the following loss term to our loss : 
\begin{align*}
    \mathcal{L} = \sum_{e \in \mathcal{S}} \|\| \alpha_{r_e} - \alpha_{r_{\sigma(e)}} \| - Iso(r_e, r_{\sigma(e)}) \|^2
\end{align*}

\subsection{Multi-Head Concatenation}

PCA analysis shows that the retained variance by embedding the isostericity matrices in lower dimensions yields the following retained variances : 0.25, 0.23, 0.17, 0.13, 0.08. - Or the cumulative version : 0.25, 0.48, 0.65, 0.78, 0.86). 
This implies that the isostericity information cannot be efficiently compacted in a scalar value.
We wish to extend the above similarity result to attention vector of a higher dimension to be able to embed the isostericity relationships in this space in a faithful way.

We can build in the same way multi-headed networks where we then concatenate the different heads' outputs.
For instance for concatenation, the norm factorization result extends to multi-headed network for norm 1 and 2. 
Therefore, we can use multi-dimensional attention weights and not be stuck with the 'embedding triangles in a line' problem.

\begin{align*}
    \mb{h}_v^{(t)} = \textrm{ReLU}\left((\mb{W}^{(t)}_{0}\mb{h}_v^{(t-1)} + \sum_{r \in R}\sum_{u \in \mathcal{N}_r(v)}\bigoplus_k \alpha^k_{r}\mb{W}\mb{h}_u^{(t-1)}\right) \\
\end{align*}

In this setting, we explicitly check that the loss we derived above, but extended in the vector setting remains relevant :

\begin{align*}
\Delta(r_1, r_2) &= \| \bigoplus_k \alpha^k_{r_1}\mb{W} \mb{h}_w^{(t-1)} - \bigoplus_k \alpha^k_{r_2} \mb{W} \mb{h}_w^{(t-1)}\|_2^2\\
&= \sum_k \sum_d (\alpha^k_{r_1} - \alpha^k_{r_2})^2 \big(\mb{W} \mb{h}_w^{(t-1)}\big)^2_d \\
&= \| \mb{\alpha}_{r_1} - \mb{\alpha}_{r_2}  \|_2^2 \| \mb{W} \mb{h}_w^{(t-1)} \|_2^2.
\end{align*}

This does not hold for all multi-head aggregation.
For instance in the summation setting, we have the following result that does not enable using our isostericity loss :

\begin{align*}
\| \sum_k \alpha^k_{r_1}\mb{W} \mb{h}_w^{(t-1)} - \sum_k \alpha^k_{r_2} \mb{W} \mb{h}_w^{(t-1)}\|_2 
= \| \sum_k \alpha^k_{r_1} - \alpha^k_{r_2} \|  \| \mb{W} \mb{h}_w^{(t-1)} \|
\neq \| \mb{\alpha}_{r_1} - \mb{\alpha}_{r_2}  \| \| \mb{W} \mb{h}_w^{(t-1)} \|.
\end{align*}

\subsection{Loss Scaling}
We have another problem, attention is not only changing the messages, it also zeroes out the irrelevant ones. We should not push such messages to have norm following the isostericity pattern. 

For this reason, we propose a twist over the loss : 
we scale each term with the normalized values of the attention vector norm, 

\begin{align*}
    \mathcal{L}_{\text{scaled}} &= \sum_{e \in \mathcal{S}}
    \mb{w}(e)
    \|\| \alpha_{r_e} - \alpha_{r_{\sigma(e)}} \| - Iso(r_e, r_{\sigma(e)}) \|^2\\
    \mb{w}(e) &= \frac{|S| \| \alpha_{r_e} \|\| \alpha_{r_{\sigma(e)}} \|}{\sum_{i \in \mathcal{S}}\| \alpha_{r_i} \|\| \alpha_{r_{\sigma(i)}} \|}.
\end{align*}

\section{Results}

We use the \texttt{rnaglib}\cite{mallet2021rnaglib} benchmark to test our layers. This benchmark tries to predict three node-level tasks : whether a node was chemically modified (a proxy for its accessibility) or binds to a protein or a small molecule. We present the results in \textbf{Table \ref{table:bench2}}.

\begin{table}[H]
\centering
\begin{tabular}{lrrr}
\toprule
\textbf{Task} : & Protein Binding & Small-molecule Binding & Chemical Modification \\
\midrule
Baseline & \textbf{0.63} & \textbf{0.60} & 0.75  \\
Iso-GCN (no scaling) & 0.59 & 0.57 & 0.75 \\
Iso-GCN (scaling) & 0.61 & 0.56 & \textbf{0.77} \\
\bottomrule
\end{tabular}
\vspace{0.1cm}
\caption{Performance of our models compared to the baseline in \textit{rnaglib}.}
\label{table:bench2}
\end{table}

We see that without scaling, the model underperform a standard RGCN. 
Adding scaling improves the results and makes the network marginally better than the baseline on the chemical modification, while marginally worse on the two other tasks.

\section{Conclusion}

This framework enables baking isostericity into graph neural network layers.
Despite the theoretical appeal of adding prior information into the network, the empirical validation does not seem to show improved results.

\bibliographystyle{unsrt}

\end{document}